# Chapter 14

# Machine Translation of Novels in the Age of Transformer

Antonio Toral is assistant professor in Language Technology at the University of Groningen. He holds a PhD in Computational Linguistics from the University of Alicante and has been researching in the field of Machine Translation (MT) since 2010. His research interests include the application of MT to literary texts, MT for under-resourced languages and the analysis of translations produced by machines and humans.

Antoni Oliver is a lecturer and the director of the Master's degree in Translation and Technologies at the Universitat Oberta de Catalunya (UOC). He holds a degree in Telecommunications Engineering, a degree in Slavonic Philology, a Master's degree in Free Software and a PhD in Linguistics. His research interests are Machine Translation, translation aid tools and automatic terminology extraction.

Pau Ribas Ballestín earned a degree in Catalan Language and Literature (UOC, 2017), and a Master's degree in Specialized Translation (UOC, 2019). He continues his relationship with the Faculty of Arts and Humanities at UOC, both as a student and as a research collaborator.

# 1.  Introduction

Machine Translation (MT) is widely used nowadays in different use-cases, including gisting and publication. While MT is applied ubiquitously, in these and other use-cases, for many text types, there is a historically perceived wisdom that MT is not suitable for literary texts.

In this chapter we build MT systems tailored to the literary domain, specifically to novels, based on the state-of-the-art architecture in neural MT (NMT), the Transformer (Vaswani et al., 2017), for the translation direction English-to-Catalan. Subsequently, we assess to what extent such a system can be useful by evaluating its translations. In so doing, this chapter builds upon the work by Toral and Way (2018), which built NMT systems adapted to novels for the aforementioned translation direction under the previous dominant paradigm: recurrent architecture with attention (Bahdanau et al., 2015).

The rest of this chapter is organized as follows. Section 2 provides an overview of previous work targeting literary texts in MT. Section 3 presents the MT systems and data sets used in this research study. The automatic and human evaluations of these systems are presented in Sections 4 and 5, respectively. Finally, Section 6 outlines conclusions and lines of future work.

# 2.  Previous Work

The existing studies on the feasibility of using computers to help translate literary texts have focused to date on three main aspects:

- Trying to detect the main characteristics of the literary style and compare these characteristics to those of the texts that are currently being translated by MT and post-editing, for example, technical and legal texts.

- Using the computer for tasks other than MT to analyses both the source and target text in order to make a better informed translation with the goal of preserving the reading experience.

- To study the use of MT to help translate literary texts through a process of MT and post-editing (MTPE).

Most previous studies about MT of literary texts are quite recent (from 2012 onwards). An important fact that has affected the conclusions from the first works to the most recent is the emergence of a relatively new paradigm of MT: Neural Machine Translation (NMT). The great increase in quality of NMT systems, compared to their predecessor, Statistical Machine Translation (SMT), has boosted the interest in MT for literary texts.

According to Toral and Way (2015), a key challenge in literary translation is preserving not only the meaning, but also the reading experience. This is a key difference to other domains, for example the above-mentioned technical or legal texts.

A recent book (Youdale, 2019) is a case study of the translation of a Spanish novel into English using technology that, in the words of the author, can lead to a stylistically better informed translation. The technology used in this case study is limited in practice to corpus linguistic tools and visualization techniques. In this approach neither MT nor computer-aided translation (CAT) tools are used. The author argues that style is important for literature because it focuses both on what is said, and on how it is said. In his opinion, tools that can help to detect or measure some stylistic characteristics would play an important role as a tool to aid translators. Youdale cites a list of common functionalities of some programs used in his approach: word frequency lists, KWIC (Key Word in Context) view, concordance displays, collocation analysis and keyword lists.

Voigt and Jurafsky (2012) demonstrate the importance of textual features above the sentence level in literary texts. They argue that human translators are able to capture the greater reference cohesion of literature, while MT systems underperform at capturing this cohesion. They suggest that, to use MT in literature, discourse features above the sentence level should be incorporated in the MT systems.

Toral and Way (2015) estimate translatability of literary texts using parallel corpora by measuring the degree of freedom of the translations (using word alignment perplexity) and the narrowness of the domain (using language model perplexity). They compare literary texts, technical documentation and news, with the overall goal of assessing the applicability of MT to literary texts.

Three features of 12 English novels are analyzed to compare the performance of SMT and NMT systems for translation into Catalan inToral and Way (2018). Namely, they analyzed lexical richness, novelty with respect to training data and average sentence length. They reported a sensitivity of NMT systems to sentence length, as reported in many other studies for different subject domains.

Some authors have also studied the usefulness of CAT tools for the translation of literary texts. The main component of these tools are the translation memories, that is, repositories of source segments with their corresponding translation to one or more languages. This aid can be of great help in text typologies (once again, such as technical or legal) with a lot of repetitions. This is not the case in literary texts, but some authors (Rothwell, 2016) have reported TMs to be useful for retranslation of classical literature, where previous translations can be retrieved automatically.

Besacier (2014) tries to provide a preliminary answer to the question: can MTPE be used to translate literary works? A short story is translated and post-edited from English to French. Then the result is revised by non-professional translators. He also presents a post-editing platform where readers can suggest post-edits of the translation. He foresees a community of reader-post-editors that will allow the continuous improvement of the translated work, making a simile with fansubs, where non-professional translators help subtitle their favorite series and movies. Use of an SMT system based on Moses trained with a 25 million-segment corpus is described. The system, however, is not tailored to literary texts. They report automatic evaluation figures (BLEU) and post-editing time for each section of the work. Human evaluation is also provided through a questionnaire to 9 readers and the views of a professional writer and translator. As a conclusion, the authors suggest that MTPE offers the possibility to have literary works translated into many languages in a 'low cost' fashion. But, is it always worth sacrificing the quality of the translation for a wider dissemination? This question remains still open.

Toral and Way (2015) also use SMT systems. The translation of a French novel into English using a freely available generic web-based MT system was evaluated. They also evaluate the translation of a Spanish novel into Catalan using an SMT system specifically trained for literary texts. They evaluate the systems using automatic metrics and use these values to select fragments with scores higher, lower and equal to the av-erage value. They performed a manual analysis of these fragments. The authors concluded with some interesting ideas for future work: adaptation of the system to a given author; treatment of cohesion and figurative language and the design of a suitable workflow and environment for MTPE.

Toral and Way (2015b) explore the use of domain adaptation techniques for an SMT system to translate a novel from Spanish to Catalan. Their results show that MT can be useful to assist with the translation of literary texts between closely-related languages.

The first reports on the comparison of SMT and NMT systems for literary translation can be found in Moorkens et al. (2018), Toral and Way (2018) and Toral, Wieling and Way (2018). In the first paper an experiment was performed with six professional translators to produce an English to Catalan literary translation under three conditions: translation from scratch, NMT post-editing and SMT post-editing. The authors reported an increase in productivity when post-editing NMT, but professional translators still stated a preference for translation from scratch as they felt less constrained and found less limitations for creativity. The authors also reported that the move from SMT to NMT has led to an improvement in quality and a decrease in the number of edit operations required. Further research is needed to find out how NMT can help both in terms of productivity and creativity. A further analysis of the results of this experiment is presented in Toral, Wieling and Way (2018). Three parameters were measured: all keystrokes, the time taken to translate each sentence as well as the number of pauses and their duration. From these data they studied the task of post-editing under three dimensions: temporal, technical and cognitive effort. The time spent is the indication of the temporal effort; the number of keystrokes the indication of the technical effort and the pauses the indication of the cognitive effort. They found that both MT approaches increase productivity, reduce the number of keystrokes and also reduce the cognitive effort. The use of MT, however, leads to longer pauses. They also reported, as in other studies, that the length of the sentence significantly affects the performance of the NMT system.

In Toral and Way (2018) tailored SMT (using Moses[1]) and NMT (using Nematus[2]) systems are trained for the translation of novels from English to Catalan. They use a large corpus of literary texts (over 100 million words). Both automatic evaluation (using BLEU) and human evaluation (perception by native readers of the target language) show that the NMT system performed substantially better.

Kuzman (2019) reports on experiments on the translation of novels from English to Slovene. They provide automatic and human evaluation figures of a tailored NMT system compared to Google Translate. Due to the relatively small training corpus for training the tailored system, the evaluation results are worse than those from Google Translate. Nevertheless, the authors report two interesting findings: a) despite the lower quality there are still gains in productivity compared to manual translation and b) the system tailored to a given author obtained promising results. This finding is in line with the suggestion from Toral (2015) and opens a door for the creation of author-tailored NMT systems by using previous translations of the same author, combined with a model trained from larger literary corpora.

Matusov (2019) presents tailored NMT systems for translating literature from English to Russian and German to English. He used the RETURNN toolkit[3] with different architectures depending on the language. For English to Russian they used an attention-based recurrent model with additive attention, whereas for German to English they used a Transformer model with multi-head attention. They report that the tailored system leads to better automatic evaluation metrics than general domain NMT. They also present a new error classification schema designed for literary machine translation. They found few syntactic errors but a significant amount of meaning errors for ambiguous words. He also analyzed and classified consistency, pronoun resolution and tone/register errors, and concluded that MT quality can improve taking into account broader contexts, as the previous sentence of the whole text.

---

1     http://www.statmt.org/moses/

2     https://github.com/EdinburghNLP/nematus

3     https://github.com/rwth-i6/returnn

In all previous works the ultimate goal of the process is to produce a high-quality translation that preserves both the content and the reading experience of the original work. Recent proposals have also centered on other possible uses of the resulting translation, such as for example, the creation of bilingual e-books as a reading aid for learners of the source language (Oliver et al., 2019). In this case the main goal is to preserve the content of the work, as the reading experience is expected to be obtained by reading the original work. In such bilingual e-books, the reader can switch between the original and the translation by clicking a given sentence or paragraph. The goal is to assist the reader in understanding difficult sentences, avoiding the need to constantly consult a bilingual dictionary.

Other interesting uses of MT in literary texts would be the translation from less popular or novel authors who have no access to the production costs of human translation and a regular publishing process. This would be an opportunity for the authors to access the international market, and, as Matusov (2019) suggests, an opportunity for readers to get to know new authors and cultures. This author also suggests that MT can help in the selection of books to be professionally translated for publication, as editors can use MT to cheaply obtain a preliminary translation and, based upon it, decide whether it is feasible to publish the book in the target language.

MT of literary texts is an area of research that is awakening the interest of a growing number of researchers, as shown in the recent workshop 'The Qualities of Machine Translation' (Hadley, 2019), held at the Machine Translation Summit XVII in Dublin (Ireland). This workshop featured 10 interesting presentations focusing on aspects ranging from productivity issues to other uses of NMT for literary works. Productivity in general is discussed in Kuzman (2019), whereas Ó Murchú (2019) focuses on time and effort. Sahin (2019) explores the use of NMT for the retranslation of classical texts. A new proposal of error evaluation tailored for literature machine translation is presented in Matusov (2019) and other works deal with error rates (Tezcan, 2019) and a comparison of quality provided by general NMT systems versus tailored NMT system for literary texts is presented in Toral (2019). Several presentations were devoted to different features of literary texts, as indirect discourse (Taivalkoski-Shilov, 2019), stylistic differences between human and neural translation (Tezcan, 2019) and translations of metaphors by humans and NMT systems (Zajdel, 2019). Regarding other uses of NMT for liter-ature, the creation of bilingual e-books is presented in Oliver et al. (2019) and the translation of Homeric classics in Sklaviadis (2019).

A related event that also took place in 2019 is the panel 'Translation technologies for creative-text translation', held at the 9th Congress of the European Society for Translation Studies in Stellenbosch (South Africa). Presentations related to the topic of this chapter included surveys of the use and attitudes to technology by translators of creative texts (Teixeira, 2019; Daems, 2019; Ruffo, 2019), the use of dictation for creative text translation (Zapata, 2019), an approach to computer-assisted literary translation focused on ecological validity (Kenny and Winters, 2019), an assessment of the reading engagement of machine-translated literary text (Guerberof-Arenas and Toral, 2019) and an account of sustainability challenges related to translation technologies when they are used for literary translation (Taivalkoski-Shilov, 2019).

# 3. Systems and Data

## 3.1. Data

The data sets used in our experiments are those previously used by Toral and Way (2018). This allows us to establish a fair comparison to the MT systems built in that previous work. The data sets fall under three types:

- Training data used to train the MT systems. This data set comprises 133 parallel novels (over 1 million sentence pairs) and around 1,000 monolingual novels in the target language (over 5 million sentences). In addition to this domain-specific data, we also use generic data sets, both parallel (0.4 million sentence pairs) and monolingual (16 million sentences).

- Development data used to evaluate the systems during training. This data set is made up of 2,000 sentence pairs randomly selected from the parallel training data and removed from it.

- Test data, which is used to evaluate the systems after training. This consists of 12 English novels and their translations by professional literary translators into Catalan. In addition to these 12 novels, in this work we also consider the short story 'Adventure II: The Yellow Face', part of the *Memoirs of Sherlock Holmes*, by Arthur Conan Doyle.

## 3.2. Systems

As previously mentioned in the introduction, we built an MT system based on the Transformer architecture (Vaswani et al., 2017). Given that the Transformer is the state-of-the-art architecture for MT, and that our system is trained on sizable amounts of in-domain data (see Section 3.1), the translations produced by our system can be considered representative of the quality that can be achieved nowadays by MT for translating novels from English to Catalan, and other translation directions that involve related languages.

In our experiments we use three additional systems. Two of these are also domain-specific (they are trained on the same data[4] as the aforementioned Transformer-based system) but use architectures that were the state of the art in the past: neural recurrent (Bahdanau et al., 2015) and statistical phrase-based (Zens et al., 2002). The third and last system, Google Translate, does use the state-of-the-art Transformer architecture but is a generic system since it is not trained on novels specifically.

This additional set of systems allows us to make two comparisons:

- Domain-specific versus generic, by confronting our Transformer-based system against Google Translate. This way we can measure the impact of the domain-specific training data on translation quality.
- MT paradigm, by confronting the Transformer-based system against the systems based on the neural recurrent and statistical phrase-based architectures. Because these systems are trained on the same data, any difference in translation quality can be attributed to the different architectures used.

---

4 With the exception of the generic training data. While the phrase-based system uses it (together with the domain-specific data by means of interpolation), the neural systems are trained solely on domain-specific training data.

# 4.    Automatic Evaluations

We automatically evaluate the translations produced by the three systems that are trained on domain-specific data (see Section 3.2) on the aforementioned set of 12 novels (see Section 3.1) with the most widely used automatic evaluation metric: BLEU (Papineni 2002).

The results are shown in Figure 1. The first neural architecture (recurrent), compared to the previous dominant architecture (phrase-based), led to an overall relative improvement of 11.4%. In turn, the Transformer brings a relative improvement of 14.5% compared to statistical phrase-based and 3.3% compared to recurrent. Finally, we evaluate the impact of fine-tuning[5] in addition to Transformer; this results in relative improvements of 22.5%, 8.3% and 5.2% compared to phrase-based, recurrent and Transformer, respectively.

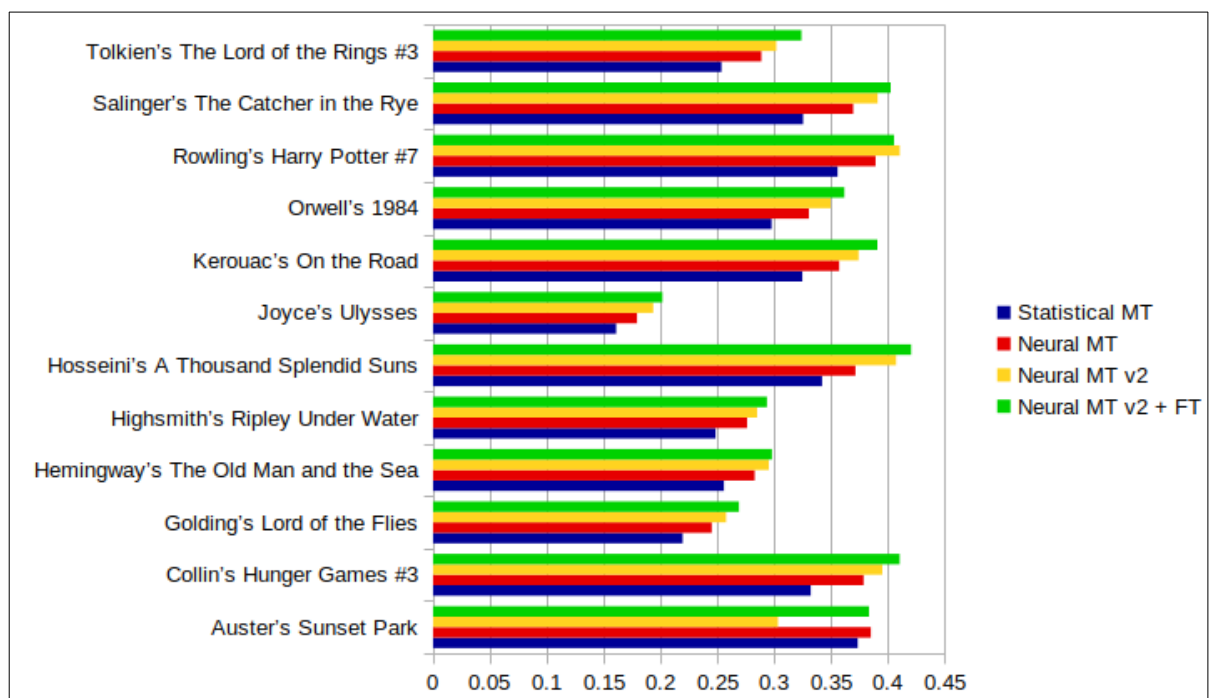

Figure 1: BLEU scores on 12 novels by the three domain-specific MT systems, and with the additional use of fine tuning (FT).

---

5    A system is typically trained on the concatenation of all the training data available (parallel and monolingual) until convergence. With fine-tuning, training is resumed after convergence with the parallel data only, again until convergence.

# 5.   Human Evaluations

While automatic evaluation metrics, such as BLEU, allow us to evaluate sizable test sets cheaply (e.g. the 12 novels evaluated in Section 4 contain over 73,000 sentences overall), they also have significant shortcomings. Therefore, we complement this automatic evaluation with appropriate human evaluations.

We conduct two types of human evaluations, to which we refer as preference and post-editing. In a nutshell, our aim with them is to answer the following questions:

- With preference: given two translations, which one is better?
- With post-editing: given a translation, how much does it need to be edited to make it 'acceptable' to the end user, a native speaker of the target language?

## 5.1.   Preference

Preference-based human evaluation is normally conducted in MT with the aim of finding out which of two (or more) translations produced by MT systems is better, given a reference human translation. Here we are not only interested in comparing translations by different MT systems, but also in comparing translations by humans and MT systems, by ranking them. Specifically, given the source sentence and two translations thereof, A and B, the task of the annotator is to rank these translations as follows: A>B if translation A is better than B, B>A if translation B is better than A, or A=B if their quality is equivalent.

Given the time-consuming nature of this task, we conduct this evaluation on subsets of three of the novels: Orwell's *1984*, Rowling's *Harry Potter #7* and Salinger's *The Catcher in the Rye*. For each novel, 10 passages of 10 consecutive sentences (100 sentences in total) were evaluated by two annotators. The annotators are professional translators who are native Catalan speakers with an advanced level of English (C2 of the Common European Framework of Reference for Languages). This is the same setup of human evaluation as Toral and Way (2018), except that in that previous work the annotators were not translators. Given that non-expert translators are more lenient toward translation errors than professional translators (Castilho et al, 2017), we expect the current evaluation to be more severe on the MT systems than that by Toral and Way (2018).

The previous human evaluation by Toral and Way (2018) confronted the human translation and the statistical phrase-based and neural recurrent systems. Because that evaluation revealed that the phrase-based system was outperformed by a large margin by the neural recurrent system for all the three books targeted, the phrase-based system is not included in the current evaluation. In other words, this evaluation considers the human translation and the neural recurrent and Transformer systems.

Figure 2 shows the results for each of the two MT systems versus the human translation. It is worth noting that the figure depicts cases where the human translation was considered better than MT (HT>MT in the figure) and where both human and MT were considered to be of equivalent quality (HT=MT) but not those annotations where MT was considered better than HT. The latter were omitted because such annotations can occur for reasons that do not entail the human translation being worse than that of the MT system, e.g. (i) a choice of the human translator to translate the sentence in a way that diverges considerably from the source sentence and (ii) a misalignment when preparing the test sets; because their sentence pairs were automatically aligned, there may be misalignments. The reader is referred to Toral and Way (2018) for a detailed account on how the test sets were built.

Back to the results shown in Figure 2: we observe that in all three books, the percentage of sentences in which the annotators perceive the translation of the MT system to be equivalent to that of the human translator is consistently higher for the Transformer-based system than for the recurrent system: 14.7% vs 9.8% for Orwell's, 31.8% vs 22.2% for Rowling's and 30.1% vs 23.4% for Salinger's.

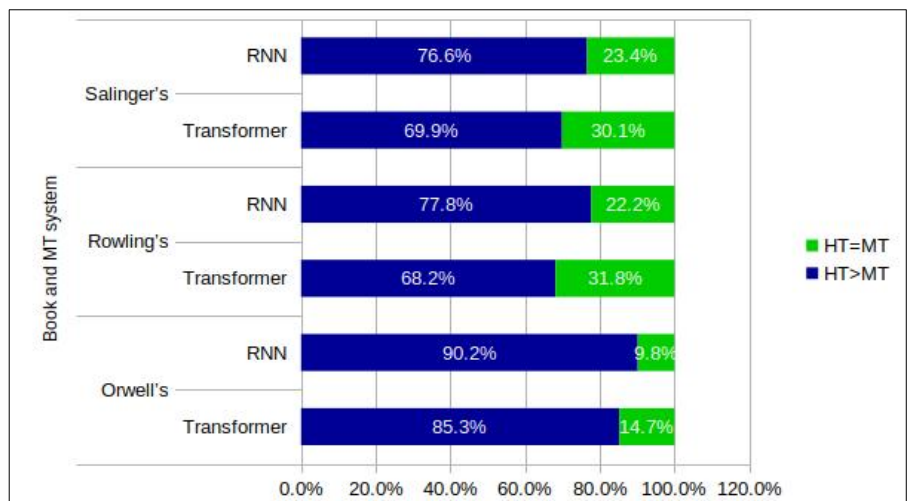

Figure 2: Results of the preference-based human evaluation for MT systems versus human translations.

Having looked at the results of the MT systems versus the human translation, we now move our attention to confronting the two MT systems against each other. The results can be found in Figure 3. In all the books the trends are similar: the biggest chunk (46.4% to 53.5%) corresponds to cases where translations by the Transformer-based system are considered better than those by the recurrent system. The second regards cases in which the translations by both systems are considered to be of equivalent quality (31.5% to 44.5%). Finally, the smallest portion (8% to 19.1%) corresponds to cases in which the translations by the recurrent system were deemed to be better than those by the Transformer-based system.

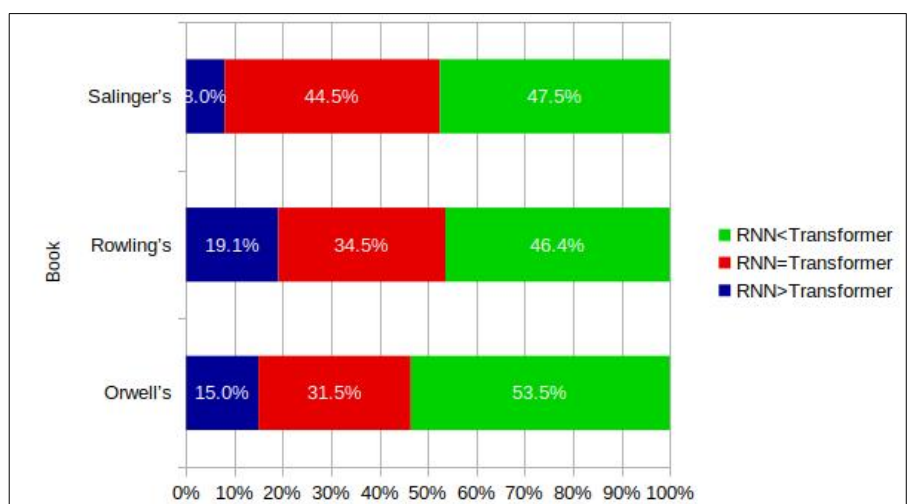

Figure 3: Results of the preference-based human evaluation for the MT systems: RNN versus Transformer.

## 5.2.  Post-editing and Error Analysis

In this section we briefly describe and show the results of a limited experiment performed on a set of random sentences from the short story 'Adventure II: The Yellow Face', part of the *Memoirs of Sherlock Holmes*, by Arthur Conan Doyle.

Our aim was, once the English to Catalan MT output had been obtained, to give the right amount of manual post-editing to each translated segment[6], in order not to fill them with literariness, as segments were decontextualized, but to make them just grammatically correct and suitable for a more thorough literary post-editing process that takes the discourse context into account.

### 5.2.1.  Segment classification

Each segment was classified[7] into six levels of required post-editing effort, according to the amount and typology of the errors, as it is shown in Table 1:

| REQ. POST-ED. | MT OUTPUT PERCEPTION |
|---|---|
| None | No post-editing required. No errors. |
| Minimal | Easy to solve, e.g., desinential and concord errors. |
| Limited | Up to one word, e.g., lexical errors. |
| Moderate | More than one word, e.g., syntactical errors. |
| Considerable | More than one word, e.g., semantic, cognitive errors. |
| Retranslation | Too much post-editing required. |

Table 1: Levels of required post-editing effort.

---

6   Post-editing was performed by a native Catalan speaker with a high level of English, linguistics and translation knowledge.

7   Although it may seem difficult to be precise, the six-level segment classification allows us to better show the distribution weighting for each MT system's output.

The MT output segments distribution is shown in Figure 4 below. A generic MT system (Google Translate) was incorporated into the experiment to establish a comparison with our customized recurrent and Transformer MT systems.

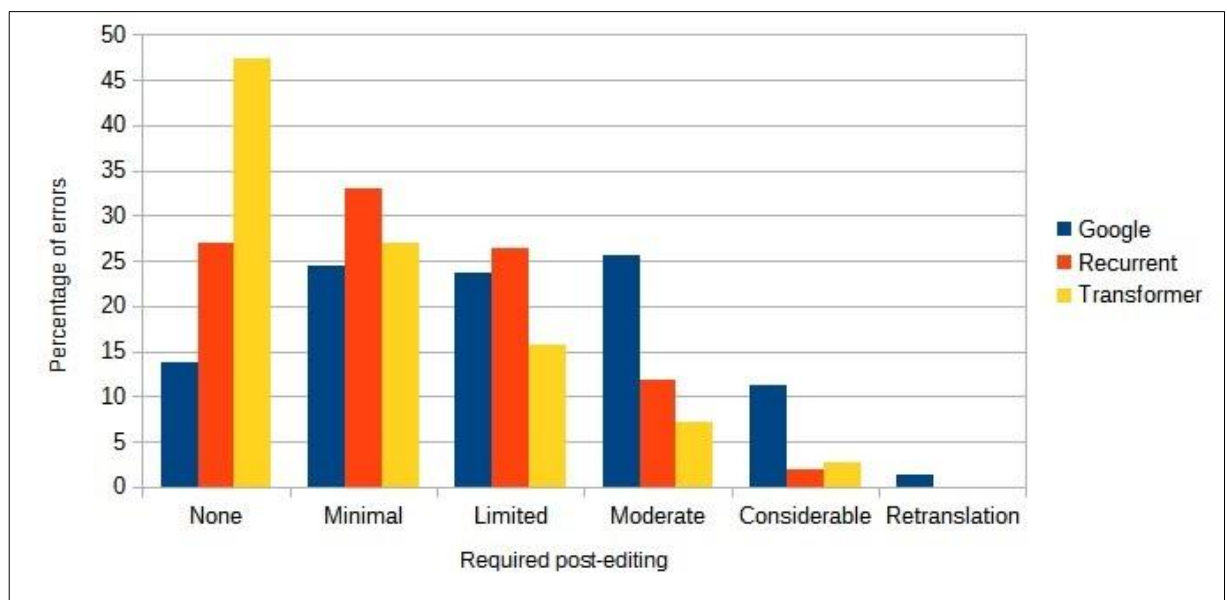

Figure 4: Percentage of segments that require post-editing for each MT system.

## 5.2.2. Error analysis

A straightforward calculation provides the average of errors per segment for each system performance, as follows: Google Translate, 1.48; Recurrent, 1.05; Transformer, 0.80.

Compared to Google Translate (see Section 3.2), the NMT Recurrent system reduces omission errors (skipping the translation of one or more consecutive words) to non-critical cases, mostly affecting only one or two consecutive words. Even though lexical errors are the most abundant, most of them affect only one word; closed grammatical categories, e.g., pronouns, prepositions, adverbs; and flexion, e.g., verbal tense, gender endings, concord (ex. 1)[8]. Usually this kind of error is closely related to the lack of context knowledge.

---

8    Examples consist of: original sentence in English, MT output in Catalan plus literal meaning in English, and minimum post-editing in Catalan.

(1)     The man must value *the pipe highly* when he prefers to patch *it* up rather than buy *a new one* with the same money.

(\*)[9] L'home ha de donar Ø valor a *la pipa* quan prefereix apedaçar-*lo* en lloc de comprar-ne *un de nou* amb els mateixos diners. [Lit. En.: The man must value the pipe(fem.) (omission: highly) when he prefers to patch it(masc.) up rather than buy a(masc.) new(masc.) one(masc.) with the same money.]

L'home ha de donar *gran valor* a *la pipa* quan prefereix apedaçar-*la* enlloc de comprar-ne *una de nova* amb els mateixos diners.

The NMT Transformer system reduces the overall amount of errors even more. Some lexical errors are still present due to polysemy. Syntactic errors are not as serious as they were in the outputs of the previous MT systems, even though some omissions are present. The only errors that require a thorough human post-editing are idiomatic expressions (ex. 2).

(2)     Mr. Grant Munro *pushed impatiently forward*, however, and we *stumbled after him* as best we could.

(?) El senyor Grant Munro *va empènyer amb impaciència cap endavant*, però, i *vam ensopegar amb ell* tan bé com vam poder. [Lit. En.: Mr. Grant Munro pushed(literal meaning) impatiently forward, however, and we bumped into him as best we could.]

El senyor Grant Munro *va avançar amb impaciència*, però, i *vam anar a trompicons darrere d'ell* tan bé com vam poder.

The high percentage of correct segments for this system (over 45%, see Figure 4) has allowed us to look for more subtle errors, which had not been considered before, as we had to deal with the more serious errors present in the previous MT systems' output. Nevertheless, the fact that a segment is correct does not always avoid a minimal post-editing call, because it may still lack a non-critical bit of information. We have classified this information loss in three types:

- ■ Person
- ■ Verbal aspect and mode
- ■ Movement direction and manner

---

9    MT output segments have been qualified with *agrammaticality* (\*), if they do not obey the rules of grammar; *questionability* (?), if their meaning is possible, but arguable.

Loss of person information creates ambiguity between action agents and patients (ex. 3):

(3)     *She* followed me, however, before *I* could close the door.

Però Ø em va seguir abans que Ø pogués tancar la porta. [Lit. En.: (Omission: she) Someone(a not specified third person) followed me, however, before (omission: I) someone(me or a not specified third person) could close the door.]

Però *ella* em va seguir abans que *jo* pogués tancar la porta.

Verbal aspect and mode information losses usually appear when translating verbal periphrasis, both aspectual (beginning, development, ending) and modal (obligation, possibility, etc.) (ex. 4):

(4)     As I did so I *happened to glance* out of one of the upper windows, and saw the maid...

En fer-ho, *vaig mirar* per una de les finestres de dalt i vaig veure la minyona… [Lit. En.: As I did so I looked out from one of the upper windows, and saw the maid…]

En fer-ho, *vaig poder llambregar* per una de les finestres de dalt i vaig veure la minyona...

Movement direction and manner losses are usually produced when translating from satellite-framed to verb-framed languages (ex. 5):

(5)     I don't know what there was about that face, Mr. Holmes, but it seemed to *send a chill right down my back*.

No sé què hi havia en aquella cara, senyor Holmes, però em va semblar que *em venia un calfred a l'esquena*. [Lit. En.: I don't know what there was about that face, Mr. Holmes, but it seemed that a chill was coming to my back.]

No sé què hi havia en aquella cara, senyor Holmes, però em va semblar que *m'enviava un calfred esquena avall*.

# 6.    Conclusions and Future Work

The aim of this chapter was to ascertain the translation quality that can currently be achieved for novels. To this end, we trained a machine translation system with sizable amounts of domain-specific parallel data (original novels and their translations) using the current state-of-the-art architecture for machine translation (Transformer). While the language pair targeted in all our experiments was English-to-Catalan, similar results should be expected for any other translation direction that involves related languages.

We evaluated our machine translation system with three evaluations. The first is automatic and uses the most-widely used automatic evaluation metric, BLEU. The two remaining evaluations are manual and they assess, respectively, preference and amount of post-editing required to make the translation error-free.

As baselines, we used (i) two systems trained on domain-specific data using architectures that were previously the state of the art (statistical phrase-based and recurrent with attention) and (ii) an off-the-shelf generic system that uses the state-of-the-art Transformer architecture (Google Translate). By so doing, we were able to evaluate the impacts on translation quality (i) of different architectures and (ii) of domain-specific versus generic training data. As expected, the Transformer-based system trained on domain-specific data outperformed the three other systems in all the three evaluations conducted, in all cases by a large margin.

For the sake of reproducibility, we make our evaluation materials available publicly, including the anonymized human evaluations. These can be found at https://github.com/antot/mt_lit_transformer/.

As for future work, we consider two research directions. The first one has to do with further improvements of the machine translation system, for example by training it on segments longer than isolated sentences (e.g. paragraphs). The second concerns possible uses of such a system, e.g. the creation of bilingual books (Oliver et al., 2019).

# Acknowledgements

The translations used for the preference-based human evaluation were ranked by Felip Turull Guerrero and by a translator who preferred to remain anonymous. We would like to thank the Center for Information Technology of the University of Groningen for their support and for providing access to the Peregrine high performance computing cluster.